\documentclass{vgtc}                          

\usepackage{times}                     

\usepackage{tabu}                      
\usepackage{booktabs}                  
\usepackage{lipsum}                    
\usepackage{mwe}                       
\usepackage{enumitem}
\usepackage{mathptmx}                  
\usepackage{dblfloatfix}  
\usepackage{multirow}
\usepackage{arydshln}
\usepackage{placeins}
\onlineid{0}

\vgtccategory{Research}

\vgtcinsertpkg

\title{AnythingReality: Robust Online Gaussian Splatting SLAM for Open-Vocabulary VR Scene Exploration}

\author{Timofei Kozlov$^{1}$\thanks{e-mail: Timofei.Kozlov@skoltech.ru} %
\and Dmitrii Maliukov$^{1}$\thanks{e-mail: Dmitrii.Maliukov@skoltech.ru} %
\and Andrey Marchenko$^{2}$\thanks{e-mail: Timofei Kozlov@skoltech.ru} %
\and Miguel Altamirano Cabrera$^{1}$\thanks{e-mail: m.altamirano@skoltech.ru} %
\and Dzmitry Tsetserukou$^{1}$\thanks{e-mail: d.tsetserukou@skoltech.ru} 
}
\affiliation{\scriptsize 
$^{1}$ Intelligent Space Robotics Laboratory, Skolkovo Institute of Science and Technology, Moscow, Moscow, Russian Federation \\
$^{2}$ NLP Research Center, Moscow, Russian Federation}

\teaser{%
  \centering
  \vspace*{-2mm}
  
  \noindent\hspace*{-0.05\textwidth}%
  \includegraphics[width=1\textwidth]{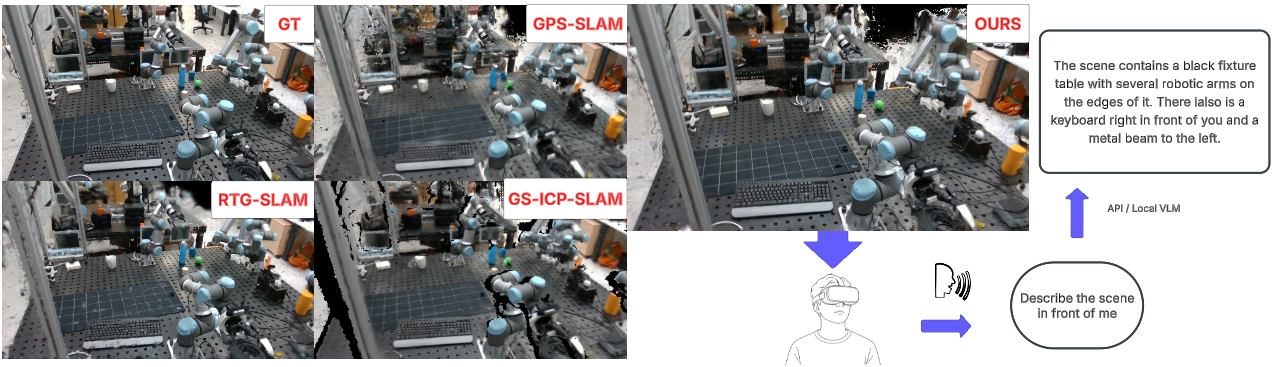}%
  \vspace{-0.3cm}
  \caption{Renders of a 3D GS reconstruction with 3 state-of-the-art methods and with our integrated system. Flow chart on the right illustrates the semantic interaction pipeline of our project.}%
  \label{fig:teaser}%
}

\abstract{

    We present a novel integrated architecture for robust online 3D Gaussian splatting, real-time VR exploration, and speech-driven Vision-Language-Model interaction. Unlike methods assuming clean depth or external poses, our system combines ORB-SLAM3-based pose estimation with online Gaussian reconstruction for noisy real-world data. A VR pipeline enables immersive exploration of incremental reconstructions; a semantic module transcribes voice commands, generates scene descriptions, and records points of interest. Against state-of-the-art online Gaussian splatting methods, we improve image quality on our dataset (+14.5\%~PSNR, +8.6\%~SSIM, $-14.3$\%~LPIPS) and TUM-RGBD (+11.7\%~PSNR, +7.8\%~SSIM, $-21.6$\%~LPIPS), with comparable or superior frame rates via quality–speed configurations. We achieve 88\% VLM object-recognition rate.

} 


\begin{document}


\firstsection{Introduction}

\maketitle

Recent advances in neural scene representations have significantly improved the quality and efficiency of 3D reconstruction. In particular, Gaussian Splatting has emerged as a powerful approach for real-time rendering of complex scenes, making it especially attractive for interactive visualization and immersive applications.

However, most self-sufficient online reconstruction pipelines require accurate depth measurements for tracking, which can be difficult to obtain in real-world settings. \cite{gpsslam} These pipelines often rely on ICP-based tracking methods, which can be sensitive to missing depth and noisy measurements of the standard RGB-D cameras. Some aspects of the inaccuracies in measurements can be avoided by using RGB-D cameras that utilize TOF sensors. However, this type of camera is significantly less budget-friendly and has problems operating under direct sunlight and is less suitable for mobile platforms in terms of energy and compute. Other online reconstruction pipelines depend on external pose estimators, such as mobile robotic systems with reliable odometry and SLAM systems. \cite{rffr}  Moreover, most reconstruction pipelines still provide limited support for direct user interaction during the reconstruction process. As a result, users often remain passive observers, with few tools for inspecting, querying, or semantically understanding the evolving scene representation.

These limitations suggest that online reconstruction should not only focus on producing accurate 3D representations, but also on making them accessible, interpretable, and interactive for human users. In this context, Virtual Reality has become an effective interface for exploring reconstructed 3D environments, as it allows users to observe spatial structure, geometry, and visual appearance from an egocentric, human-scale perspective \cite{lobachev,koessler,rodriguez}. This immersive form of inspection is particularly useful when reconstructed scenes need to be assessed not only as geometric models, but also as spaces that can be navigated, interpreted, and analyzed by humans. At the same time, recent advances in vision-language models (VLMs) have shown that visual content can be effectively linked to natural-language descriptions, questions, and instructions, enabling more flexible forms of semantic reasoning \cite{radford, liu}. Building on this progress, extending these capabilities to 3D scenes has enabled language-guided object localization, question answering, scene understanding, and interactive reasoning in reconstructed environments \cite{chen, azuma, hong}. Recent work further explores VLMs for 3D scene understanding from visual context \cite{qi2026gpt4scene} and language-aware Gaussian representations for open-vocabulary segmentation and object localization \cite{peng2025visionlanguagegs}.

\begin{figure}[h    ]
  \centering
  \includegraphics[width=0.85\linewidth]{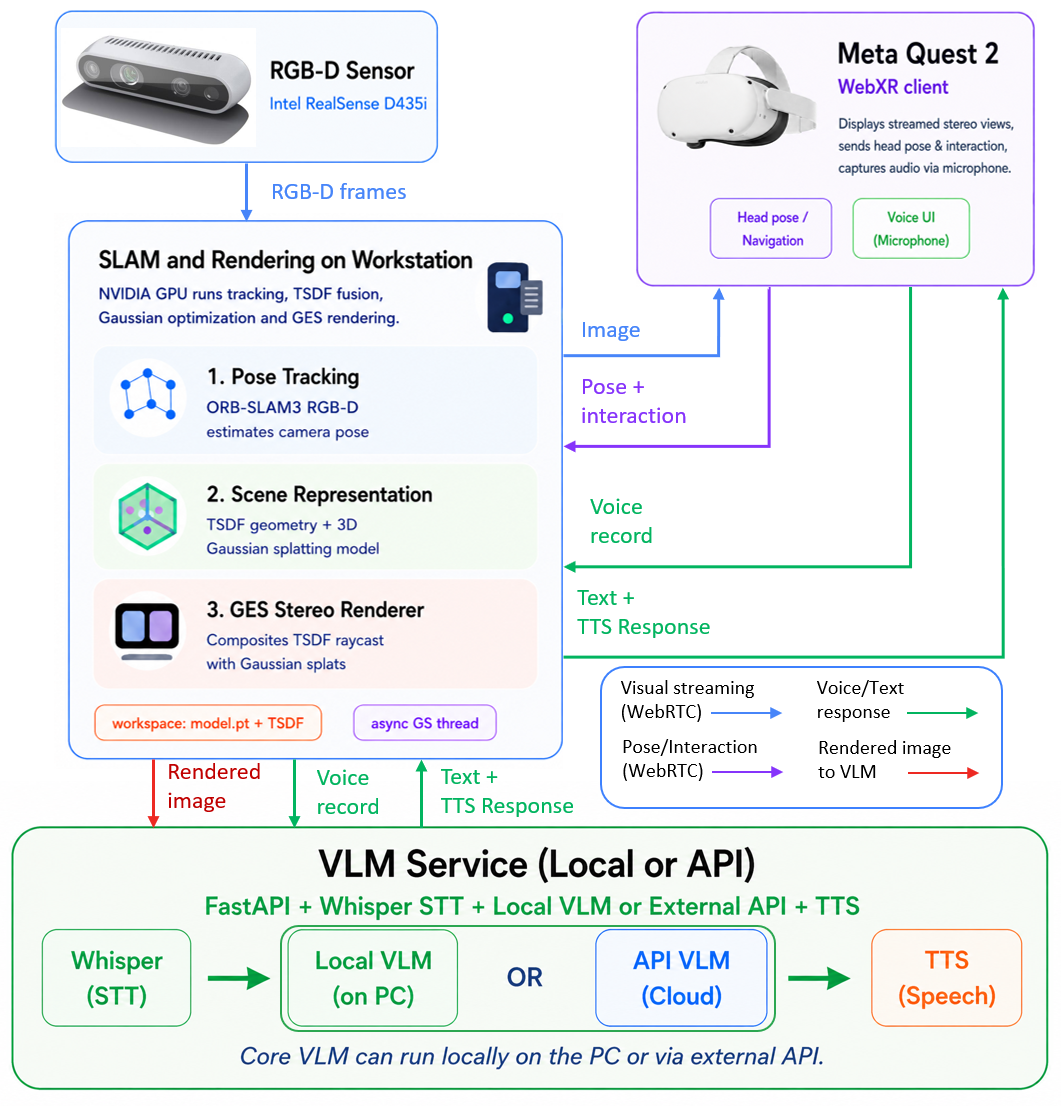}
  \vspace{-0.5cm}
  \caption{ Overview of the AnythingReality pipeline: online mapping, stereo VR streaming, and speech-driven VLM interaction.}
  \vspace{-0.6cm}
  \label{fig:gsvr-rendering}
\end{figure}

In this work, we present an integrated system for online 3D Gaussian Splatting reconstruction, real-time VR exploration, and speech-driven VLM-assisted interaction. The system uses ORB-SLAM-based camera pose estimation to reduce dependence on clean depth-based tracking and integrates the resulting pose stream with an online Gaussian reconstruction module, a VR rendering pipeline, and a semantic interaction module. Users can explore the scene during reconstruction, ask natural-language questions about the current Gaussian-rendered view, and create open-vocabulary points of interest by voice. We evaluate the system on several indoor scenes and publicly available datasets, measuring both online reconstruction frame rate and rendering quality using PSNR, SSIM, and LPIPS. The results demonstrate the system’s potential for immersive scene understanding, remote inspection, and human-in-the-loop reconstruction.

\section{System Architecture}

Our system couples four modules: (i) RGB-D capture and pose estimation with a lightweight ORB-SLAM3 backend, (ii) online Gaussian-plus-SDF mapping, (iii) stereo VR rendering and WebXR streaming of the incrementally reconstructed map, and (iv) VLM-based speech-driven semantic interaction.

\subsection{Online Tracking and Reconstruction}

We build our online mapping module on the Gaussian-plus-SDF SLAM framework \cite{gpsslam}, which combines fast volumetric RGB-D fusion with selective 3D Gaussian optimization. In that design, a truncated signed distance field (TSDF) provides coarse geometry and initial appearance, while a sparse set of 3D Gaussians refines regions where SDF rendering
deviates from observed color. We adopt this hybrid representation as our \emph{mapping core}, but expand it to a streaming and deployable online system for RGB-D sensors with noisy depth measurements. We replace batch dataset loading with a frame-by-frame live input path (RealSense capture or deterministic replay for benchmarking), decouple Gaussian optimization from capture through an asynchronous thread when needed, and expose the incrementally updated map to a stereo VR renderer over WebRTC. 

We further improve the original spawn policy for noisy indoor depth: in addition to color-error-driven insertion, we gate new Gaussians by raycast depth and TSDF integration confidence. Each incoming RGB-D frame is processed causally. Given the current camera pose, the system integrates color and depth into the global TSDF, raycasts the fused volume to obtain SDF color and depth, and compares the result against the live image. Pixels with a large photometric error and sufficient geometric reliability become candidates for Gaussian insertion; a subset is sampled and initialized on the raycast surface. Every $K$ frames, the system runs a short burst of GES optimization over a sliding window of recent views and selected keyframes, then prunes unstable Gaussians. The updated Gaussian map is immediately available for evaluation metrics and for VR rendering, without waiting for an offline training pass over the full sequence.

For desktop evaluation we reuse the GES forward pass from the mapping core. For VR telepresence we render two eye views from the user's head pose taking into account the distance between the eye lenses of the headset and stream JPEG to the client. Thus the user explores the \emph{same} incrementally built representation that the mapper updates online, rather than a post-processed model produced after scanning completes.

\subsection{GS-VR Rendering}

The GS-VR rendering module provides an online stereo visualization pipeline for immersive VR scene exploration (Fig.~\ref{fig:gsvr-rendering}). Recent Gaussian-based VR systems demonstrate that 3DGS can support interactive immersive manipulation and rendering of reconstructed content \cite{jiang2024vrgs}; our renderer focuses on streaming the incrementally reconstructed online map for inspection during SLAM. Instead of transferring the full Gaussian map to the headset for every frame, the system performs high-quality GES rendering on the host PC and streams the resulting stereo views to the VR client. At each WebXR frame, the headset sends its current head pose to the SLAM server, where the pose is aligned with the SLAM coordinate frame and used to generate left and right eye views from the current online Gaussian scene representation.

For each eye, the renderer constructs a virtual camera using the configured IPD (interpupillary distance), field of view, and headset pose. The scene is rendered using the same Gaussian-plus-SDF pipeline used for desktop visualization. The resulting stereo images are encoded and transmitted through a low-latency WebRTC channel to the headset.

On the VR side, the received frames are decoded and displayed as per-eye textures in WebXR. To reduce perceived rotational latency, the client applies lightweight asynchronous timewarp: between two server-rendered frames, the displayed image is reprojected according to the latest headset rotation. This allows the user to perceive smooth head rotation at the headset refresh rate, while the computationally expensive Gaussian rendering remains online on the PC. Translation and joystick-based navigation still require newly rendered frames from the server, but the system prioritizes frame freshness over frame buffering to keep interaction responsive during live scene exploration.

\begin{table*}[!t]
  \caption{Comparison of runtime performance and rendering quality on the Replica, TUM-RGBD, and our four RealSense sequences (Our\_Lab, Our\_Objects, Our\_Tables, Our\_Kitchen). Our method (AnythingReality, AR)is presented with two configurations: first one focuses on reconstruction quality, the second focuses on maximizing the frame rate. Best and second-best results in each column are shown in \textbf{bold} and \underline{underlined}, respectively.}
  \label{tab:runtime_rendering_quality}
  \footnotesize%
  \centering%
  \begin{tabular}{llcccccc}
  \toprule
  \textbf{Method} & \textbf{Metric} & \textbf{Replica} & \textbf{TUM-RGBD} & \textbf{Our\_Lab} & \textbf{Our\_Objects} & \textbf{Our\_Tables} & \textbf{Our\_Kitchen} \\
  \midrule

  \multirow{4}{*}{RTG-SLAM}
  & FPS$\uparrow$       & 14.6  & 16.3  & 5.1  & 6.4  & 4.6  & 5.8 \\
  & PSNR$\uparrow$      & 22.82 & 12.13 & 17.61 & 20.7 & 22.0 & 18.5 \\
  & SSIM$\uparrow$      & 0.751 & 0.454 & 0.571 & 0.692 & 0.788 & 0.738 \\
  & LPIPS$\downarrow$   & 0.459 & 0.437 & 0.521 & 0.428 & 0.398 & 0.458 \\

  \hdashline

  \multirow{4}{*}{GS-ICP SLAM}
  & FPS$\uparrow$       & \underline{202.1} & 108.14 & \underline{110.2} & 114.55 & 80.3 & 89.7 \\
  & PSNR$\uparrow$      & \textbf{40.07} & \underline{17.23} & 19.45 & 21.59 & 24.42 & 21.45 \\
  & SSIM$\uparrow$      & \textbf{0.976} & \underline{0.694} & 0.648 & 0.754 & 0.886 & 0.834 \\
  & LPIPS$\downarrow$   & \textbf{0.048} & \underline{0.365} & 0.405 & 0.369 & \underline{0.315} & 0.427 \\

  \hdashline

  \multirow{4}{*}{GPS-SLAM}
  & FPS$\uparrow$       & \textbf{208.8} & \textbf{202.8} & 73.8 & \textbf{175.8} & \textbf{239.4} & \underline{101.89} \\
  & PSNR$\uparrow$      & \underline{39.95} & 12.52 & 19.69 & 24.68 & 24.78 & 19.68 \\
  & SSIM$\uparrow$      & \underline{0.968} & 0.407 & 0.624 & 0.787 & 0.877 & 0.813 \\
  & LPIPS$\downarrow$   & \underline{0.106} & 0.541 & 0.455 & 0.365 & 0.335 & 0.395 \\

  \hdashline

  \multirow{4}{*}{\begin{tabular}[c]{@{}l@{}}AnythingReality (Quality)\\(ours)\end{tabular}}
  & FPS$\uparrow$       & 61.2 & 89.4 & 47.2 & 58.3 & 59.1 & 63.2 \\
  & PSNR$\uparrow$      & \underline{37.84} & \textbf{19.26} & \textbf{22.69} & \textbf{27.72} & \textbf{27.02} & \textbf{23.91} \\
  & SSIM$\uparrow$      & \underline{0.958} & \textbf{0.747} & \textbf{0.724} & \textbf{0.851} & \textbf{0.907} & \textbf{0.869} \\
  & LPIPS$\downarrow$   & \underline{0.111} & \textbf{0.286} & \textbf{0.335} & \textbf{0.278} & \textbf{0.303} & \textbf{0.347} \\

  \hdashline

  \multirow{4}{*}{\begin{tabular}[c]{@{}l@{}}AnythingReality (FPS)\\(ours)\end{tabular}}
  & FPS$\uparrow$       & 170.4 & \underline{201.7} & \textbf{141.5} & \underline{175.3} & \underline{178.3} & \textbf{223.8} \\
  & PSNR$\uparrow$      & 34.66 & 15.04 & \underline{21.57} & \underline{26.68} & \underline{26.11} & \underline{23.46} \\
  & SSIM$\uparrow$      & 0.935 & 0.645 & \underline{0.686} & \underline{0.834} & \underline{0.897} & \underline{0.866} \\
  & LPIPS$\downarrow$   & 0.166 & 0.385 & \underline{0.379} & \underline{0.304} & 0.318 & \underline{0.353} \\

  \bottomrule
  \end{tabular}%
  \vspace{-0.6cm}
\end{table*}

\subsection{Speech-Driven Semantic Interaction}

The semantic interaction module provides a voice-based interface for querying and annotating the reconstructed scene while the user remains immersed in VR. The VR client sends a push-to-talk audio command to the interaction server together with the current headset pose and, for visual questions, the current Gaussian-rendered view. The audio is transcribed with a local faster-whisper model~\cite{radford2023whisper}. Whisper-style speech recognition is suitable for this interface because it supports multilingual short-form commands and robust transcription across diverse acoustic conditions, while server-side inference keeps speech processing close to the VR pipeline. The faster-whisper implementation is used as a practical low-latency inference backend with GPU and CPU deployment options.

The VLM component is exposed through an OpenAI-compatible service interface. This keeps the interaction layer independent from a specific model provider: the same API can be configured for hosted inference or for local serving through vLLM-style deployments~\cite{kwon2023efficient}. In our prototype, the VLM is used in two complementary roles. A text model handles frequent command routing and label extraction, while a vision-language model is invoked for image-conditioned questions about the current Gaussian-rendered view.

The transcript is first processed by a text VLM that classifies the command into one of three intents: marking a point of interest, describing the current view, or rejecting unsupported commands. To make the interaction predictable, the router uses a narrow intent schema with the labels \textit{mark}, \textit{describe}, and \textit{unsupported}. The service requests structured JSON outputs, validates model responses against application schemas, and uses fallback structured-output modes when needed. This lets downstream VR logic consume typed responses instead of free-form text.

For point-of-interest marking, the system extracts an open-vocabulary label from commands such as ``mark this as a table'' or ``label this as point one''. The server returns this label together with the pose or coordinate supplied by the VR client at the moment of marking. The resulting coordinate-label point of interest gives users a simple way to annotate meaningful locations during online reconstruction. Future work can extend these records toward object-level grounding and persistent 3D semantic anchoring.

For scene description, the module sends the user's question and the current Gaussian-rendered image to the VLM. The generated answer is returned as text for the VR interface. This view-based interaction gives the user an immediate and lightweight way to ask questions about the visible reconstructed scene during immersive inspection. Future extensions can combine such responses with speech feedback and broader map-level semantic memory.

Unsupported commands, such as navigation requests, deletion commands, greetings, or unclear speech, are handled explicitly. The server returns a short message explaining that the current interaction modes are limited to scene description and point-of-interest marking. This prevents accidental scene changes and keeps the user interface predictable during reconstruction.

\section{System Evaluation}

We evaluated the system using two groups of metrics: reconstruction performance, measured by online reconstruction frame rate and rendering quality in terms of PSNR, SSIM, and LPIPS, and semantic understanding, measured by the object-level recognition rate of the VLM in the reconstructed scenes.

\subsection{Reconstruction evaluation}

To assess the reconstruction part of our pipeline, we recorded a dataset of 6,000 frames, consisting of four RGB-D videos captured in different indoor scenes with varying object density, average camera-to-object distance, and lighting conditions. All recordings were made with an Intel RealSense D435i RGB-D camera, which uses a stereo-vision-based depth sensor. As a result, the depth images contained occlusion regions caused by stereo matching limitations, which became more extreme as the distance between the camera and the observed objects decreased. In addition, the recorded depth maps exhibited significant noise and artifacts.

To ensure that our system can provide online high-quality reconstructions for VR inspection, we compared the performance of our solution to the state-of-the-art Gaussian RGB-D SLAM methods such as GS-ICP SLAM, RTG-SLAM and GPS-SLAM (see Table 1). We tested all systems on our dataset as well as open-access datasets Replica and TUM-RGBD. To evaluate the performance of our reconstruction subsystem, the replay pipeline was used, allowing us to imitate the video stream from the RealSense camera. 

As shown in Table 1, our system outperforms the compared methods in terms of image quality on all four of our recorded scenes, which contain noisy depth maps and occlusion regions. This indicates improved robustness under imperfect depth observations. Notably, our method resulted in the least Gaussian count at each of our four scenes, along with the TUM-RGBD dataset, compared to other methods (mean 47\% decrease in the number of Gaussians compared to GPS-SLAM, which uses SDF fusion to limit Gaussian growth). This is a result of the several spawn gating conditions and filters created by us specifically for noisy datasets with uncertain depth measurements. The state-of-the-art methods' configurations and settings were modified for this comparison to account for a camera that is different from the original authors'. Without those changes, these methods achieved significantly lower results.

\subsection{Semantic Interaction Evaluation}

We evaluated the semantic interaction module using two types of voice commands: scene-description queries and point-of-interest marking commands. For description queries, the user asked natural-language questions about the currently visible reconstructed view, and the VLM generated a text response based on the Gaussian-rendered image. For marking commands, the user issued instructions such as ``mark this as a chair'' or ``label this as inspection point one'', and the system extracted an open-vocabulary label together with the current VR pose or coordinate.

To measure VLM object-recognition, we sampled rendered views from the reconstructed indoor scenes and manually annotated whether the main visible objects could be recognized from the rendered image. We then asked the VLM to describe the current view or identify visible objects. A response was considered successful if it mentioned the dominant visible object or produced a semantically correct description of the scene region. Using this protocol, the VLM achieved an understandable-view recognition rate of 88\%. 

For point-of-interest marking, the command router reliably distinguished marking commands from description requests and unsupported commands. The structured-output interface reduced parsing ambiguity by forcing the model response into a small set of supported intents: \textit{mark}, \textit{describe}, and \textit{unsupported}. In successful marking cases, the extracted labels were stored together with the corresponding 3D coordinate or headset pose, enabling lightweight semantic annotation during online exploration.

The semantic evaluation shows that even a view-based VLM interface can provide useful interaction with an online Gaussian reconstruction. However, the current module reasons only over the rendered image observed by the user and does not yet maintain a persistent object-level semantic map. Therefore, semantic annotations are stored as open-vocabulary point records rather than fully grounded 3D object instances. Extending the system with persistent object segmentation, multi-view language grounding, and map-level semantic memory remains a direction for future work.

\begin{figure}[t]
  \centering
  \includegraphics[width=0.7\linewidth]{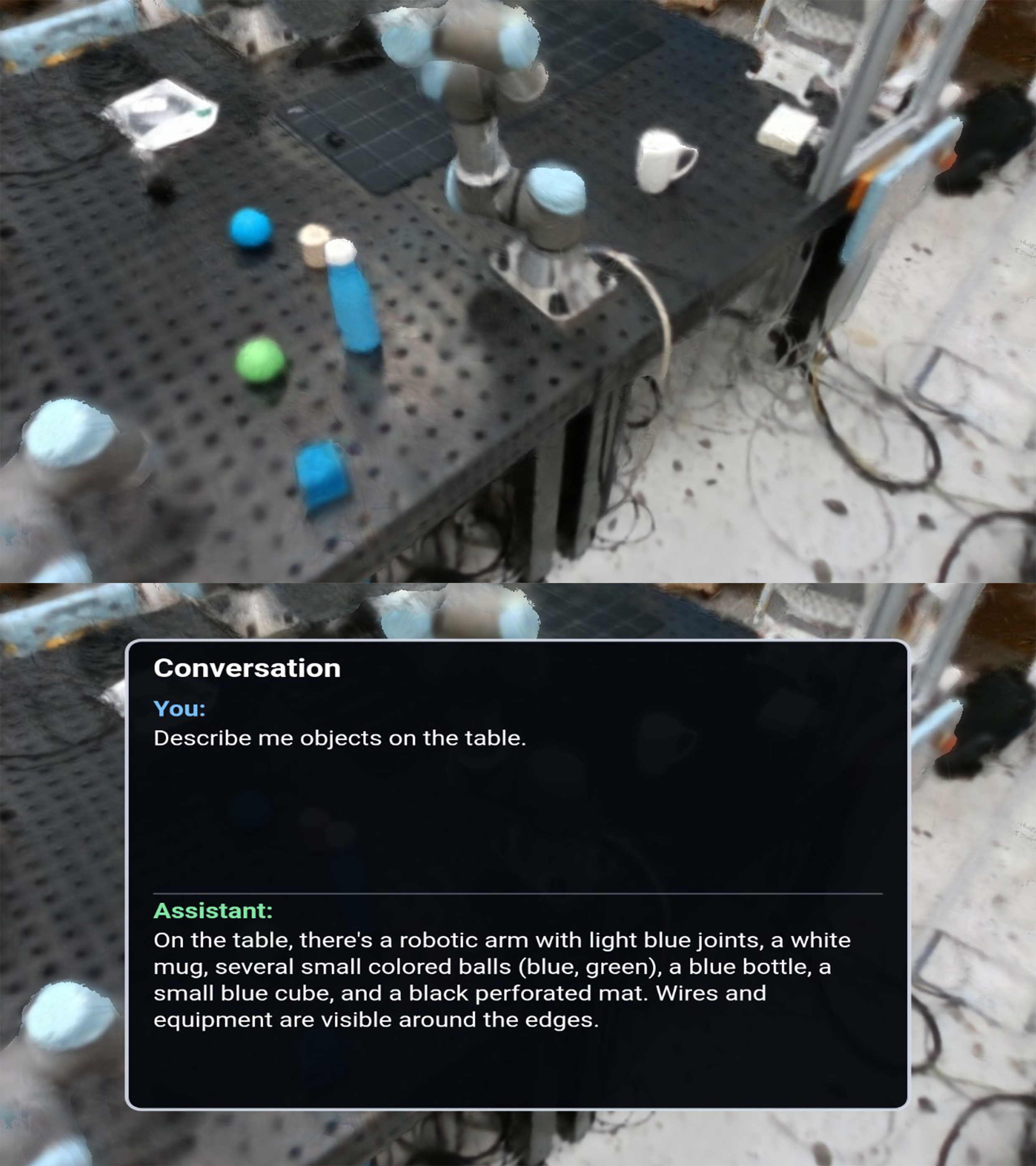}
  \caption{
  Example of semantic interaction in VR. 
  The upper view shows the Gaussian-rendered tabletop scene observed by the user, 
  while the lower overlay shows the dialogue interface where a natural-language 
  scene-description query is answered by the VLM.
  }
  \vspace{-0.7cm}
  \label{fig:semantic-interaction}
\end{figure}

\section{Conclusion}

We developed AnythingReality, a complete system built for robust online 3D Gaussian Splatting reconstruction, immersive VR scene exploration, and speech-driven semantic interaction. The proposed architecture integrates RGB-D input processing, ORB-SLAM3-based pose estimation, online Gaussian-plus-SDF mapping, stereo VR rendering, WebRTC streaming, and VLM-assisted interaction into a single online pipeline.

Our system allows users to inspect an incrementally reconstructed scene directly in VR, rather than waiting for an offline reconstruction stage. By relying on SLAM-based pose estimation and online Gaussian mapping, AnythingReality improves robustness under noisy and incomplete RGB-D observations. At the same time, the semantic interaction module enables natural-language scene queries and open-vocabulary point-of-interest annotation, making the reconstruction process more interactive and human-centered.

The developed system demonstrates the potential of combining Gaussian Splatting, SLAM, VR, and vision-language models for immersive inspection, remote operation, and robotic perception. Future work will focus on persistent semantic mapping, multi-view object grounding, improved 3D annotation anchoring, and further optimization of reconstruction and streaming latency.



\end{document}